# Learning Factor Graphs in Polynomial Time & Sample Complexity

**Pieter Abbeel**
Computer Science Dept.
Stanford University
Stanford, CA 94305

**Daphne Koller**
Computer Science Dept.
Stanford University
Stanford, CA 94305

**Andrew Y. Ng**
Computer Science Dept.
Stanford University
Stanford, CA 94305

## Abstract

We study computational and sample complexity of parameter and structure learning in graphical models. Our main result shows that the class of factor graphs with bounded factor size and bounded connectivity can be learned in polynomial time and polynomial number of samples, assuming that the data is generated by a network in this class. This result covers both parameter estimation for a known network structure and structure learning. It implies as a corollary that we can learn factor graphs for both Bayesian networks and Markov networks of bounded degree, in polynomial time and sample complexity. Unlike maximum likelihood estimation, our method does not require inference in the underlying network, and so applies to networks where inference is intractable. We also show that the error of our learned model degrades gracefully when the generating distribution is not a member of the target class of networks.

## 1 Introduction

Graphical models are widely used to compactly represent structured probability distributions over (large) sets of random variables. The task of learning a graphical model representation for a distribution $P$ from samples taken from $P$ is an important one for many applications. There are many variants of this learning problem, which vary on several axes, including whether the data is fully or partially observed, and on whether the structure of the network is given or needs to be learned from data.

In this paper, we focus on the problem of learning both network structure and parameters from fully observable data, restricting attention to discrete probability distributions over finite sets. We focus on the problem of learning a factor graph representation of the distribution (Kschischang et al., 2001). Factor graphs subsume both Bayesian networks and Markov networks, in that every Bayesian network or Markov network can be written as a factor graph of (essentially) the same size.

Based on the canonical parameterization used in the Hammersley-Clifford Theorem for Markov networks (Hammersley & Clifford, 1971; Besag, 1974), we provide a parameterization of factor graph distributions that is a product only of probabilities over local subsets of variables. By contrast, the original Hammersley-Clifford canonical parameterization is a product of probabilities over joint instantiations of all the variables. The new parameterization naturally leads to an algorithm that solves the parameter estimation problem in closed-form. For factor graphs of bounded factor size and bounded connectivity, if the generating distribution falls into the target class, we show that our estimation procedure returns an accurate solution — one of low KL-divergence to the true distribution — given a *polynomial number of samples*.

Building on this result, we provide an algorithm for learning both the structure and parameters of such factor graphs. The algorithm uses empirical entropy estimates to select an approximate Markov blanket for each variable, and then uses the parameter estimation algorithm to estimate parameters and identify which factors are likely to be irrelevant. Under the same assumptions as above, we prove that this algorithm also has polynomial-time computational complexity and polynomial sample complexity.[1]

These algorithms provide the first polynomial-time and polynomial sample-complexity learning algorithm for factor graphs, and thereby for Markov networks. Note that our algorithms apply to any factor graph of bounded factor size and bounded connectivity, including those (such as grids) where inference is intractable. We also show that our algorithms degrade gracefully, in that they return reasonable answers even when the underlying distribution does not come exactly from the target class of networks. We note that the proposed algorithms are unlikely to be useful in practice in their current form, as they do an exhaustive enumeration on the possible Markov blankets of factors in the factor graph, a process which is generally infeasible even in small networks; they also do not make good use of all the available data. Nevertheless, the techniques used in our analysis opens new avenues towards efficient parameter and structure learning in undirected, intractable models.

---

[1] Due to space constraints, most the proofs are omitted from this paper or given only as sketches. The complete proofs are given in the full paper (Abbeel et al., 2005).

## 2 Preliminaries

### 2.1 Factor Graph Distributions

**Definition 1** (Gibbs distribution). *A* factor *$f$ with* scope[2] *$\mathbf{D}$ is a function from $\text{val}(\mathbf{D})$ to $\mathbb{R}^+$. A Gibbs distribution $P$ over a set of random variables $\mathcal{X} = \{X_1, \ldots, X_n\}$ is associated with a set of factors $\{f_j\}_{j=1}^J$ with scopes $\{\mathbf{C}_j\}_{j=1}^J$, such that*

$$P(X_1, \ldots, X_n) = \tfrac{1}{Z} \prod_{j=1}^J f_j(\mathbf{C}_j[X_1, \ldots, X_n]). \quad (1)$$

*The normalizing constant $Z$ is the* partition function.

The *factor graph* associated with a Gibbs distribution is a bipartite graph whose nodes correspond to variables and factors, with an edge between a variable $X$ and a factor $f_j$ if the scope of $f_j$ contains $X$. There is one-to-one correspondence between factor graphs and the sets of scopes. A Gibbs distribution also induces a Markov network — an undirected graph whose nodes correspond to the random variables $\mathcal{X}$ and where there is an edge between two variables if there is a factor in which they both participate. The set of scopes uniquely determines the structure of the Markov network, but several different sets of scopes can result in the same Markov network. For example, a fully connected Markov network can correspond both to a Gibbs distribution with a factor which is a joint distribution over $\mathcal{X}$, and to a distribution with $\binom{n}{2}$ factors over pairs of variables. We will use the more precise factor graph representation in this paper. Our results are easily translated into results for Markov networks.

**Definition 2** (Markov blanket). *Let a set of scopes $\mathcal{C} = \{\mathbf{C}_j\}_{j=1}^J$ be given. The* Markov blanket *of a set of random variables $\mathbf{D} \subseteq \mathcal{X}$ is defined as*

$$\text{MB}(\mathbf{D}) = \cup \{\mathbf{C}_j : \mathbf{C}_j \in \mathcal{C}, \ \mathbf{C}_j \cap \mathbf{D} \neq \emptyset\} - \mathbf{D}.$$

For any Gibbs distribution, we have, for any $\mathbf{D}$, that

$$\mathbf{D} \perp \mathcal{X} - \mathbf{D} - \text{MB}(\mathbf{D}) \mid \text{MB}(\mathbf{D}), \quad (2)$$

or in words: given its Markov blanket, $\mathbf{D}$ is independent of all other variables.

A standard assumption for a Gibbs distribution, which is critical for identifying its structure (see Lauritzen, 1996, Ch. 3), is that the distribution be *positive* — all of its entries be non-zero. Our results use a quantitative measure for how positive $P$ is. Let $\gamma = \min_{\mathbf{x},i} P(X_i = x_i | \mathcal{X}_{-i} = \mathbf{x}_{-i})$, where the $-i$ subscript denotes all entries but entry $i$. Note that, if we have a fixed bound on the number of factors in which a variable can participate, and a bound on how skewed each factor is (in terms of the ratio of its lowest and highest entries), we are guaranteed a bound on $\gamma$ that is independent of the number $n$ of variables in the network. By contrast, $\tilde{\gamma} = \min_{\mathbf{x}} P(\mathbf{X} = \mathbf{x})$ generally has an exponential dependence on $n$. For example, if we have $n$ IID Bernoulli($\tfrac{1}{2}$) random variables, then $\gamma = \tfrac{1}{2}$ (independent of $n$) but $\tilde{\gamma} = \tfrac{1}{2^n}$.

### 2.2 The Canonical Parameterization

A Gibbs distribution is generally over-parameterized relative to the structure of the underlying factor graph, in that a continuum of possible parameterizations over the graph can all encode the same distribution. The *canonical parameterization* (Hammersley & Clifford, 1971; Besag, 1974) provides one specific choice of parameterization for a Gibbs distribution, with some nice properties (see below). The canonical parameterization forms the basis for the Hammersley-Clifford theorem, which asserts that any distribution that satisfies the independence assumptions encoded by a Markov network can be represented as a Gibbs distribution over the cliques in that network. Standardly, the canonical parameterization is defined for Gibbs distributions over Markov networks at the clique level. We utilize a more refined parameterization, defined at the factor level; results at the clique level are trivial corollaries.

The canonical parameterization is defined relative to an arbitrary (but fixed) assignment $\bar{\mathbf{x}} = (\bar{x}_1, \ldots, \bar{x}_n)$. Let any subset of variables $\mathbf{D} = \langle X_{i_1}, \ldots, X_{i_{|\mathbf{D}|}} \rangle$, and any assignment $\mathbf{d} = \langle x_{i_1}, \ldots, x_{i_{|\mathbf{D}|}} \rangle$ be given. Let any $\mathbf{U} \subseteq \mathbf{D}$ be given. We define $\sigma.[\cdot]$ such that for all $i \in \{1, \ldots, n\}$:

$$(\sigma_{\mathbf{U}}[\mathbf{d}])_i = \begin{cases} x_i & \text{if } X_i \in \mathbf{U}, \\ \bar{x}_i & \text{if } X_i \notin \mathbf{U}. \end{cases}$$

In words, $\sigma_{\mathbf{U}}[\mathbf{d}]$ keeps the assignments to the variables in $\mathbf{U}$ as specified in $\mathbf{d}$, and augments it to form a full assignment using the default values in $\bar{\mathbf{x}}$. Note that the assignments to variables outside $\mathbf{U}$ are always ignored, and replaced with their default values. Thus, the scope of $\sigma_{\mathbf{U}}[\cdot]$ is always $\mathbf{U}$.

Let $P$ be a positive Gibbs distribution. The *canonical factor* for $\mathbf{D} \subseteq \mathcal{X}$ is defined as follows:

$$f^*_{\mathbf{D}}(\mathbf{d}) = \exp\left(\sum_{\mathbf{U} \subseteq \mathbf{D}} (-1)^{|\mathbf{D}-\mathbf{U}|} \log P(\sigma_{\mathbf{U}}[\mathbf{d}])\right). \quad (3)$$

The sum is over all subsets of $\mathbf{D}$, including $\mathbf{D}$ itself and the empty set $\emptyset$.

The following theorem extends the Hammersley-Clifford theorem (which applies to Markov networks) to factor graphs.

**Theorem 1.** *Let $P$ be a positive Gibbs distribution with factor scopes $\{\mathbf{C}_j\}_{j=1}^J$. Let $\{\mathbf{C}^*_j\}_{j=1}^{J^*} = \cup_{j=1}^J 2^{\mathbf{C}_j} - \emptyset$ (where $2^{\mathbf{X}}$ is the power set of $\mathbf{X}$ — the set of all of its subsets). Then*

$$P(\mathbf{x}) = P(\bar{\mathbf{x}}) \prod_{j=1}^{J^*} f^*_{\mathbf{C}^*_j}(\mathbf{c}^*_j),$$

---
[2] A function has *scope* $\mathbf{X}$ if its domain is $\text{val}(\mathbf{X})$.

where $\mathbf{c}_j^*$ is the instantiation of $\mathbf{C}_j^*$ consistent with $\mathbf{x}$.

The parameterization of $P$ using the canonical factors $\{f_{\mathbf{C}_j^*}^*\}_{j=1}^{J^*}$ is called the *canonical parameterization* of $P$. Although typically $J^* > J$, the additional factors are all subfactors of the original factors. Note that first transforming a factor graph into a Markov network and then applying the Hammersley-Clifford theorem to the Markov network generally results in a significantly less sparse canonical parameterization than the canonical parameterization from Theorem 1.

## 3 Parameter Estimation

### 3.1 Markov Blanket Canonical Factors

Considering the definition of the canonical parameters, we note that all of the terms in Eqn. (3) can be estimated from empirical data using simple counts, without requiring inference over the network. Thus, it appears that we can use the canonical parameterization as the basis for our parameter estimation algorithm. However, as written, this estimation process is statistically infeasible, as the terms in Eqn. (3) are probabilities over full instantiations of all variables, which can never be estimated from a reasonable number of samples.

Our first observation is that we can obtain exactly the same answer by considering probabilities over much smaller instantiations — those corresponding to a factor and its Markov blanket. Let $\mathbf{D} = \langle X_{i_1}, \ldots, X_{i_{|\mathbf{D}|}} \rangle$ be any subset of variables, and $\mathbf{d} = \langle x_{i_1}, \ldots, x_{i_{|\mathbf{D}|}} \rangle$ be any assignment to $\mathbf{D}$. For any $\mathbf{U} \subseteq \mathbf{D}$, we define $\sigma_{\mathbf{U}:\mathbf{D}}[\mathbf{d}]$ to be the restriction of the full instantiation $\sigma_{\mathbf{U}}[\mathbf{d}]$ of all variables in $\mathcal{X}$ to the corresponding instantiation of the subset $\mathbf{D}$. In other words, $\sigma_{\mathbf{U}:\mathbf{D}}[\mathbf{d}]$ keeps the assignments to the variables in $\mathbf{U}$ as specified in $\mathbf{d}$, and changes the assignment to the variables in $\mathbf{D} - \mathbf{U}$ to the default values in $\bar{\mathbf{x}}$. Let $\mathbf{D} \subseteq \mathcal{X}$ and $\mathbf{Y} \subseteq \mathcal{X} - \mathbf{D}$. Then the factor $f_{\mathbf{D}|\mathbf{Y}}^*$ over the variables in $\mathbf{D}$ is defined as follows:

$$f_{\mathbf{D}|\mathbf{Y}}^*(\mathbf{d}) = \exp\left(\sum_{\mathbf{U} \subseteq \mathbf{D}} (-1)^{|\mathbf{D}-\mathbf{U}|} \log P(\sigma_{\mathbf{U}:\mathbf{D}}[\mathbf{d}]|\mathbf{Y} = \bar{\mathbf{y}})\right), \quad (4)$$

where the sum is over all subsets of $\mathbf{D}$, including $\mathbf{D}$ itself and the empty set $\emptyset$.

The following proposition shows an equivalence between the factors computed using Eqn. (3) and Eqn. (4).

**Proposition 2.** *Let $P$ be a positive Gibbs distribution with factor scopes $\{\mathbf{C}_j\}_{j=1}^J$, and $\{\mathbf{C}_j^*\}_{j=1}^{J^*}$ as above. Then for any $\mathbf{D} \subseteq \mathcal{X}$, we have:*

$$f_{\mathbf{D}}^* = f_{\mathbf{D}|\mathcal{X}-\mathbf{D}}^* = f_{\mathbf{D}|\text{MB}(\mathbf{D})}^*, \quad (5)$$

*and (as a direct consequence)*

$$P(\mathbf{x}) = P(\bar{\mathbf{x}}) \prod_{j=1}^{J^*} f_{\mathbf{C}_j^*|\mathcal{X}-\mathbf{C}_j^*}^*(\mathbf{c}_j^*) \quad (6)$$

$$= P(\bar{\mathbf{x}}) \prod_{j=1}^{J^*} f_{\mathbf{C}_j^*|\text{MB}(\mathbf{C}_j^*)}^*(\mathbf{c}_j^*), \quad (7)$$

*where $\mathbf{c}_j^*$ is the instantiation of $\mathbf{C}_j^*$ consistent with $\mathbf{x}$.*

Proposition 2 shows that we can compute the canonical parameterization factors using probabilities over factor scopes and their Markov blankets only. From a sample complexity point of view, this is a significant improvement over the standard definition which uses joint instantiations over all variables. By expanding the Markov blanket canonical factors in Proposition 2 using Eqn. (4) we see that any factor graph distribution can be parameterized as a product of local probabilities.

Table 1: Notational conventions.

| | |
|---|---|
| $X, Y, \ldots$ | random variables |
| $x, y, \ldots$ | instantiations of the random variables |
| $\mathbf{X}, \mathbf{Y}, \ldots$ | sets of random variables |
| $\mathbf{x}, \mathbf{y}, \ldots$ | instantiations of sets of random variables |
| $\text{val}(X)$ | set of values the variable $X$ can take |
| $\mathbf{D}[\mathbf{x}]$ | instantiation of $\mathbf{D}$ consistent with $\mathbf{x}$ (abbreviated as $\mathbf{d}$ when no ambiguity is possible) |
| $f$ | factor |
| $P$ | positive Gibbs distribution over a set of random variables $\mathcal{X} = \langle X_1, \ldots, X_n \rangle$ |
| $\{f_j\}_{j=1}^J$ | factors of $P$ |
| $\{\mathbf{C}_j\}_{j=1}^J$ | scopes of factors of $P$ |
| $\hat{P}$ | empirical (sample) distribution |
| $\tilde{P}$ | distribution returned by learning algorithm |
| $f^*$ | canonical factor as defined in Eqn. (3) |
| $f_{\cdot|\cdot}^*$ | canonical factor as defined in Eqn. (4) |
| $\hat{f}_{\cdot|\cdot}^*$ | canonical factor as defined in Eqn. (4), but using the empirical distribution $\hat{P}$ |
| $\text{MB}(\mathbf{D})$ | Markov blanket of $\mathbf{D}$ |
| $k$ | $\max_j |\mathbf{C}_j|$ |
| $\gamma$ | $\min_{\mathbf{x},i} P(X_i = x_i | \mathcal{X}_{-i} = \mathbf{x}_{-i})$ |
| $v$ | $\max_i |\text{val}(X_i)|$ |
| $b$ | $\max_j |\text{MB}(\mathbf{C}_j)|$ |

### 3.2 Parameter Estimation Algorithm

Based on the parameterization above, we propose the following Factor-Graph-Parameter-Learn algorithm. The algorithm takes as inputs: the scopes of the factors $\{\mathbf{C}_j\}_{j=1}^J$, samples $\{\mathbf{x}^{(i)}\}_{i=1}^m$, a baseline instantiation $\bar{\mathbf{x}}$. Then for $\{\mathbf{C}_j^*\}_{j=1}^{J^*}$ as above, Factor-Graph-Parameter-Learn does the following:

- Compute the estimates of the canonical factors $\{\hat{f}_{\mathbf{C}_j^*|\text{MB}(\mathbf{C}_j^*)}^*\}_{j=1}^{J^*}$ as in Eqn. (4), but using the empirical estimates based on the data samples.

- Return the probability distribution $\tilde{P}(\mathbf{x}) \propto \prod_{j=1}^{J^*} \hat{f}_{\mathbf{C}_j^*|\text{MB}(\mathbf{C}_j^*)}^*(\mathbf{c}_j^*)$.

**Theorem 3** (Parameter learning: computational complexity). *The running time of the Factor-Graph-Parameter-Learn algorithm is in $O(m2^k J(k+b) + 2^{2k} J v^k)$.*[3]

---
[3] The upper bound is based on a very naive implementation's running time. It assumes operations (such as reading, writing, adding, etc.) numbers take constant time.

Note the representation of the factor graph distribution is $\Omega(Jv^k)$, thus exponential dependence on $k$ is unavoidable for any algorithm. There is no dependence on the running time of evaluating the partition function. On the other hand, evaluating the likelihood requires evaluating the partition function (which is different for different parameter values). We expect that ML-based learning algorithms would take at least as long as evaluating the partition function.

### 3.3 Sample Complexity

We now analyze the sample complexity of the Factor-Graph-Parameter-Learn algorithm, showing that it returns a distribution that is a good approximation of the true distribution when given only a "small" number of samples.

**Theorem 4** (Parameter learning: sample complexity). *Let any $\epsilon, \delta > 0$ be given. Let $\{\mathbf{x}^{(i)}\}_{i=1}^m$ be IID samples from $P$. Let $\tilde{P}$ be the probability distribution returned by* Factor-Graph-Parameter-Learn. *Then, we have that, for*

$$D(P\|\tilde{P}) + D(\tilde{P}\|P) \leq J\epsilon$$

*to hold with probability at least $1 - \delta$, it suffices that*

$$m \geq (1 + \tfrac{\epsilon}{2^{2k+2}})^2 \, \tfrac{2^{4k+3}}{\gamma^{k+b}\epsilon^2} \log \tfrac{2^{k+2}Jv^{k+b}}{\delta}. \qquad (8)$$

*Proof sketch.* Using the Hoeffding inequality and the fact that the probabilities are bounded away from 0, we first prove that for any $j \in \{1, \ldots, J^*\}$, for

$$\left|\log P(\mathbf{C}_j^*|\mathrm{MB}(\mathbf{C}_j^*)) - \log \hat{P}(\mathbf{C}_j^*|\mathrm{MB}(\mathbf{C}_j^*))\right| \leq \epsilon' \quad (9)$$

to hold with high probability, a "small" number of samples is sufficient. Now using the fact that the logs of the (Markov blanket) canonical factors are sums of at most $2^{|\mathbf{C}_j^*|} \leq 2^k$ of these log probabilities, we get that the resulting estimates of the (Markov blanket) canonical factors are accurate:

$$|\log f^*_{\mathbf{C}_j^*|\mathrm{MB}(\mathbf{C}_j^*)}(\mathbf{c}_j^*) - \log \hat{f}^*_{\mathbf{C}_j^*|\mathrm{MB}(\mathbf{C}_j^*)}(\mathbf{c}_j^*)| \leq 2^k \epsilon'. \tag{10}$$

From Proposition 2 we have that the true distribution can be written as a product of its (Markov blanket) canonical factors. I.e., we have $P \propto \prod_{j=1}^{J^*} f^*_{\mathbf{C}_j^*|\mathrm{MB}(\mathbf{C}_j^*)}$. So Eqn. (10) shows we have an accurate estimate of all the factors of the distribution. We use a technical lemma to show that this implies that the two distributions are close in KL-divergence:

$$D(P\|\tilde{P}) + D(\tilde{P}\|P) \leq 2J^*2^k\epsilon' = J^*2^{k+1}\epsilon'. \tag{11}$$

Now using $J^* \leq J2^k$, appropriately choosing $\epsilon'$ and careful bookkeeping on the number of samples required and the high-probability statements gives the theorem. □

The theorem shows that—assuming the true distribution $P$ factors according to the given structure—Factor-Graph-Parameter-Learn returns a distribution that is $J\epsilon$-close in KL-divergence. The sample complexity scales exponentially in the maximum number of variables per factor $k$, and polynomially in $\frac{1}{\epsilon}, \frac{1}{\gamma}$.

The error in the KL-divergence grows linearly with $J$. This is a consequence of the fact that the number of terms in the distributions is $J$, and each can accrue an error. We can obtain a more refined analysis if we eliminate this dependence by considering the normalized KL-divergence $\frac{1}{J}D(P\|\tilde{P})$. We return to this issue in Section 3.4.

Theorem 4 considers the case when $P$ actually factors according to the given structure. The following theorem shows that our error degrades gracefully even if the samples are generated by a distribution $Q$ that does not factor according to the given structure.

**Theorem 5** (Parameter learning: graceful degradation). *Let any $\epsilon, \delta > 0$ be given. Let $\{\mathbf{x}^{(i)}\}_{i=1}^m$ be IID samples from a distribution $Q$. Let $\mathrm{MB}$ and $\widehat{\mathrm{MB}}$ be the Markov blankets according to the distribution $Q$ and the given structure respectively. Let $\{f^*_{\mathbf{D}_j^*|\mathrm{MB}(\mathbf{D}_j^*)}\}_{j=1}^{\tilde{J}}$ be the Markov blanket canonical factors of $Q$. Let $\{\mathbf{C}_j^*\}_{j=1}^{J^*}$ be the scopes of the canonical factors for the given structure Let $\tilde{P}$ be the probability distribution returned by* Factor-Graph-Parameter-Learn. *Then we have that for*

$$D(Q\|\tilde{P}) + D(\tilde{P}\|Q) \leq J\epsilon + 2\sum_{j: \mathbf{D}_j^* \notin \{\mathbf{C}_k^*\}_{k=1}^{J^*}} \max_{\mathbf{d}_j^*} \left|\log f^*_{\mathbf{D}_j^*}(\mathbf{d}_j^*)\right|$$

$$+ 2\sum_{j \in S} \max_{\mathbf{d}_j^*} \left|\log \frac{f^*_{\mathbf{D}_j^*|\mathrm{MB}(\mathbf{D}_j^*)}(\mathbf{d}_j^*)}{f^*_{\mathbf{D}_j^*|\widehat{\mathrm{MB}}(\mathbf{D}_j^*)}(\mathbf{d}_j^*)}\right| \qquad (12)$$

*to hold with probability at least $1 - \delta$, it suffices that $m$ satisfies Eqn. (8) of Theorem 4. Here the elements of $S = \{j \,:\, \mathbf{D}_j \in \{\mathbf{C}_k^*\}_{k=1}^{J^*}, \mathrm{MB}(\mathbf{D}_j^*) \neq \widehat{\mathrm{MB}}(\mathbf{D}_j^*)\}$ index over the canonical factors for which the Markov blanket is incorrect in the given structure.*

This result is important, as it shows that each canonical factor that is incorrectly captured by our target structure adds at most a constant to our bound on the KL-divergence. A canonical factor could be incorrectly captured when the corresponding factor scope is not included in the given structure. Canonical factors are designed so that a factor over a set of variables captures only the residual interactions between the variables in its scope, once all interactions between its subsets have been accounted for in other factors. Thus, canonical factors over large scopes are often close to the trivial all-ones factor in practice. Therefore, if our structure approximation is such that it only ignores some of the larger-scope factors, the error in the approximation may be quite limited. A canonical factor could also be incorrectly captured when the given structure does not have the correct Markov blanket for that factor. The resulting error depends on how good an approximation of the Markov blanket we do have. See Section 4 for more details on this topic.

### 3.4 Reducing the Dependence on Network Size

Our previous analysis showed a linear dependence on the number of factors $J$ in the network. In a sense, this dependence is inevitable. To understand why, consider a distribution $P$ defined by a set of $n$ independent Bernoulli random variables $X_1, \ldots, X_n$, each with parameter 0.5. Assume that $Q$ is an approximation to $P$, where the $X_i$ are still independent, but have parameter 0.4999. Intuitively, a Bernoulli(0.4999) distribution is a very good estimate of a Bernoulli(0.5); thus, for most applications, $Q$ can safely be considered to be a very good estimate of $P$. However, the KL-divergence between $D(P(X_{1:n}) \| Q(X_{1:n})) = \sum_{i=1}^{n} D(P(X_i) \| Q(X_i)) = \Omega(n)$. Thus, if $n$ is large, the KL divergence between $P$ and $Q$ would be large, even though $Q$ is a good estimate for $P$. To remove such unintuitive scaling effects when studying the dependence on the number of variables, we can consider instead the normalized KL divergence criterion:

$$D_n(P(X_{1:n}) \| Q(X_{1:n})) = \tfrac{1}{n} D(P(X_{1:n}) \| Q(X_{1:n})).$$

As we now show, with a slight modification to the algorithm, we can achieve a bound of $\epsilon$ for our normalized KL-divergence while eliminating the logarithmic dependence on $J$ in our sample complexity bound. Specifically, we can modify our algorithm so that it clips probability estimates $\in [0, \gamma^{k+b})$ to $\gamma^{k+b}$. Note that this change can only improve the estimates, as the true probabilities which we are trying to estimate are never in the interval $[0, \gamma^{k+b})$.[4]

For this slightly modified version of the algorithm, the following theorem shows the dependence on the size of the network is $O(1)$, which is tighter than the logarithmic dependence shown in Theorem 4.

**Theorem 6** (Parameter learning: size of the network). *Let any $\epsilon, \delta > 0$ be given. Let $\{\mathbf{x}^{(i)}\}_{i=1}^{m}$ be IID samples from $P$. Let the domain size of each variable be fixed. Let the number of factors a variable can participate in be fixed. Let $\tilde{P}$ be the probability distribution returned by* Factor-Graph-Parameter-Learn. *Then we have that, for*

$$D_n(P \| \tilde{P}) + D_n(\tilde{P} \| P) \leq \epsilon$$

*to hold with probability at least $1 - \delta$, it suffices that we have a certain number of samples that does not depend on the number of variables in the network.*

The following theorem shows a similar result for Bayesian networks, namely that for a fixed bound on the number of parents per node, the sample complexity dependence on the size of the network is $O(1)$.[5]

---

[4]This solution assumes that $\gamma$ is known. If not, we can use a clipping threshold as a function of the number of samples seen. This technique is used by Dasgupta (1997) to derive sample complexity bounds for learning fixed structure Bayesian networks.

[5]Complete proofs for Theorems 6 and 7 (and all other results in this paper) are given in the full paper Abbeel et al. (2005). In

**Theorem 7.** *Let any $\epsilon > 0$ and $\delta > 0$ be given. Let any Bayesian network (BN) structure over $n$ variables with at most $k$ parents per variable be given. Let $P$ be a probability distribution that factors over the BN. Let $\tilde{P}$ denote the probability distribution obtained by fitting the conditional probability tables (CPT) entries via maximum likelihood and then clipping each CPT entry to the interval $[\frac{\epsilon}{4}, 1 - \frac{\epsilon}{4}]$. Then we have that for*

$$D_n(P \| \tilde{P}) \leq \epsilon, \quad (13)$$

*to hold with probability at least $1 - \delta$, it suffices that we have a number of samples that does not depend on the number of variables in the network.*

## 4 Structure Learning

The algorithm described in the previous section uses the known network to establish a Markov blanket for each factor. This Markov blanket is then used to effectively estimate the canonical parameters from empirical data. In this section, we show how we can build on this algorithm to perform structure learning, by first identifying (from the data) an approximate Markov blanket for each candidate factor, and then using this approximate Markov blanket to compute the parameters of that factor from a "small" number of samples.

### 4.1 Identifying Markov Blankets

In the parameter learning results, the Markov blanket $\mathrm{MB}(\mathbf{C}_j^*)$ is used to efficiently estimate the conditional probability $P(\mathbf{C}_j^* | \mathcal{X} - \mathbf{C}_j^*)$, which is equal to $P(\mathbf{C}_j^* | \mathrm{MB}(\mathbf{C}_j^*))$. This suggests to measure the quality of a candidate Markov blanket $\mathbf{Y}$ by how well $P(\mathbf{C}_j^* | \mathbf{Y})$ approximates $P(\mathbf{C}_j^* | \mathcal{X} - \mathbf{C}_j^*)$. In this section we show how conditional entropy can be used to find a candidate Markov blanket that gives a good approximation for this conditional probability. One intuition why conditional entropy has the desired property, is that it corresponds to the log-loss of predicting $\mathbf{C}_j^*$ given the candidate Markov blanket.

**Definition 3** (Conditional Entropy.). *Let $P$ be a probability distribution over over $\mathbf{X}, \mathbf{Y}$. Then the conditional entropy $H(\mathbf{X}|\mathbf{Y})$ of $\mathbf{X}$ given $\mathbf{Y}$ is defined as*

$$-\sum_{\mathbf{x} \in \mathrm{val}(\mathbf{X}), \mathbf{y} \in \mathrm{val}(\mathbf{Y})} P(\mathbf{X} = \mathbf{x}, \mathbf{Y} = \mathbf{y}) \log P(\mathbf{X} = \mathbf{x} | \mathbf{Y} = \mathbf{y}).$$

**Proposition 8** (Cover & Thomas, 1991). *Let $P$ be a probability distribution over $\mathbf{X}, \mathbf{Y}, \mathbf{Z}$. Then we have $H(\mathbf{X}|\mathbf{Y}, \mathbf{Z}) \leq H(\mathbf{X}|\mathbf{Y})$.*

Proposition 8 shows that conditional entropy can be used to find the Markov blanket for a given set of variables.

---

the full paper we actually give a much stronger version of Theorem 7, including dependencies of $m$ on $\epsilon, \delta, k$ and a graceful degradation result.

Let $\mathbf{D}, \mathbf{Y} \subseteq \mathcal{X}, \mathbf{D} \cap \mathbf{Y} = \emptyset$, then we have

$$H(\mathbf{D}|\mathrm{MB}(\mathbf{D})) = H(\mathbf{D}|\mathcal{X} - \mathbf{D}) \leq H(\mathbf{D}|\mathbf{Y}), \quad (14)$$

where the equality follows from the Markov blanket property stated in Eqn. (2) and the inequality follows from Prop. 8. Thus, we can select as our candidate Markov blanket for $\mathbf{D}$ the set $\mathbf{Y}$ which minimizes $H(\mathbf{D}|\mathbf{Y})$.

Our first difficulty is that, when learning from data, we do not have the true distribution, and hence the exact conditional entropies are unknown. The following lemma shows that the conditional entropy can be efficiently estimated from samples.

**Lemma 9.** *Let $P$ be a probability distribution over $\mathbf{X}, \mathbf{Y}$ such that for all instantiations $\mathbf{x}, \mathbf{y}$ we have $P(\mathbf{X} = \mathbf{x}, \mathbf{Y} = \mathbf{y}) \geq \lambda$. Let $\widehat{H}$ be the conditional entropy computed based upon $m$ IID samples from $P$. Then for*

$$\left| H(\mathbf{X}|\mathbf{Y}) - \widehat{H}(\mathbf{X}|\mathbf{Y}) \right| \leq \epsilon$$

*to hold with probability $1 - \delta$, it suffices that*

$$m \geq \tfrac{8|\mathrm{val}(\mathbf{X})|^2|\mathrm{val}(\mathbf{Y})|^2}{\lambda^2 \epsilon^2} \log \tfrac{4|\mathrm{val}(\mathbf{X})||\mathrm{val}(\mathbf{Y})|}{\delta}.$$

However, as the empirical estimates of the conditional entropy are noisy, the true Markov blanket is *not* guaranteed to achieve the minimum of $H(\mathbf{D} \mid \mathbf{Y})$. In fact, in some probability distributions, many sets of variables could be arbitrarily close to reaching equality in Eqn. (14). Thus, in many cases, our procedure will not recover the actual Markov blanket based on a finite number of samples. Fortunately, as we show in the next lemma, any set of variables $\mathbf{Y}$ that is close to achieving equality in Eqn. (14) gives an accurate approximation $P(\mathbf{C}_j|\mathbf{Y})$ of the probabilities $P(\mathbf{C}_j|\mathcal{X} - \mathbf{C}_j)$ used in the canonical parameterization.

**Lemma 10.** *Let any $\epsilon > 0$ be given. Let $P$ be a distribution over disjoint sets of random variables $\mathbf{V}, \mathbf{W}, \mathbf{X}, \mathbf{Y}$. Let $\lambda_1 = \min_{\mathbf{v} \in \mathrm{val}(\mathbf{V}), \mathbf{w} \in \mathrm{val}(\mathbf{W})} P(\mathbf{v}, \mathbf{w})$, $\lambda_2 = \min_{\mathbf{x} \in \mathrm{val}(\mathbf{X}), \mathbf{v} \in \mathrm{val}(\mathbf{V}), \mathbf{w} \in \mathrm{val}(\mathbf{W})} P(\mathbf{x}|\mathbf{v}, \mathbf{w})$. Assume the following holds*

$$\mathbf{X} \perp \mathbf{Y}, \mathbf{W} \mid \mathbf{V}, \quad (15)$$
$$H(\mathbf{X}|\mathbf{W}) \leq H(\mathbf{X}|\mathbf{V}, \mathbf{W}, \mathbf{Y}) + \epsilon. \quad (16)$$

*Then we have that $\forall \mathbf{x}, \mathbf{y}, \mathbf{v}, \mathbf{w}$*

$$\left| \log P(\mathbf{x}|\mathbf{v}, \mathbf{w}, \mathbf{y}) - \log P(\mathbf{x}|\mathbf{w}) \right| \leq \tfrac{\sqrt{2\epsilon}}{\lambda_2 \sqrt{\lambda_1}}. \quad (17)$$

In other words, if a set of variables $\mathbf{W}$ looks like a Markov blanket for $\mathbf{X}$, as evaluated by the conditional entropy $H(\mathbf{X}|\mathbf{W})$, then the conditional distribution $P(\mathbf{X}|\mathbf{W})$ must be close to the conditional distribution $P(\mathbf{X}|\mathcal{X} - \mathbf{X})$. Thus, it suffices to find such an approximate Markov blanket $\mathbf{W}$ as a substitute for knowing the true Markov blanket. This makes conditional entropy suitable for structure learning.

### 4.2 Structure Learning Algorithm

We propose the following Factor-Graph-Structure-Learn algorithm. The algorithm receives as input: samples $\{\mathbf{x}^{(i)}\}_{i=1}^m$, the maximum number of variables per factor $k$, the maximum number of variables per Markov blanket for a factor $b$, a base instantiation $\bar{\mathbf{x}}$. Let $\mathcal{C}$ be the set of candidate factor scopes, let $\mathcal{Y}$ be the set of candidate Markov blankets. I.e., we have

$$\mathcal{C} = \{\mathbf{C}_j^* : \mathbf{C}_j^* \subseteq \mathcal{X}, \mathbf{C}_j^* \neq \emptyset, |\mathbf{C}_j^*| \leq k\}, \quad (18)$$
$$\mathcal{Y} = \{\mathbf{Y} : \mathbf{Y} \subseteq \mathcal{X}, |\mathbf{Y}| \leq b\}. \quad (19)$$

The algorithm does the following:

- $\forall \mathbf{C}_j^* \in \mathcal{C}$, compute the best candidate Markov blanket: $\widehat{\mathrm{MB}}(\mathbf{C}_j^*) = \arg\min_{\mathbf{Y} \in \mathcal{Y}, \mathbf{Y} \cap \mathbf{C}_j^* = \emptyset} \widehat{H}(\mathbf{C}_j^*|\mathbf{Y})$.

- $\forall \mathbf{C}_j^* \in \mathcal{C}$, compute the estimates $\{\hat{f}_{\mathbf{C}_j^*|\widehat{\mathrm{MB}}(\mathbf{C}_j^*)}^*\}_i$ of the canonical factors as defined in Eqn. (4) using the empirical distribution.

- Threshold to one the factor entries $\hat{f}_{\mathbf{C}_j^*|\widehat{\mathrm{MB}}(\mathbf{C}_j^*)}^*(\mathbf{c}_j^*)$ satisfying $|\log \hat{f}_{\mathbf{C}_j^*|\widehat{\mathrm{MB}}(\mathbf{C}_j^*)}^*(\mathbf{c}_j^*)| \leq \tfrac{\epsilon}{2^{k+2}}$, and discard the resulting trivial factors that have all entries equal to one.

- Return the probability distribution $\tilde{P}(\mathbf{x}) \propto \prod_i \hat{f}_{\mathbf{C}_j^*|\widehat{\mathrm{MB}}(\mathbf{C}_j^*)}^*(\mathbf{c}_j^*)$.

The thresholding step finds the factors that actually contribute to the distribution. The specific threshold is chosen to suit the proof of Theorem 12. If no thresholding were applied, the error in Eqn. (20) would be $\tfrac{|\mathcal{C}|}{2^k}\epsilon$ instead of $J\epsilon$, which is much larger in case the true distribution has a relatively small number of factors.

**Theorem 11** (Structure learning: computational complexity). *The running time of Factor-Graph-Structure-Learn is in $O\left(mkn^k bn^b(k+b) + kn^k bn^b v^{k+b} + kn^k 2^k v^k\right)$.*[6]

The first two terms come from going through the data and computing the empirical conditional entropies. Since the algorithm considers all combinations of candidate factors and Markov blankets, we have an exponential dependence on the maximum scope size $k$ and the maximum Markov blanket size $b$. The last term comes from computing the Markov blanket canonical factors. The important fact about this result is that, unlike for (exact) ML approaches, the running time does not depend on the tractability of inference in the underlying true distribution, nor the recovered structure.

**Theorem 12** (Structure learning: sample complexity). *Let any $\epsilon, \delta > 0$ be given. Let $\tilde{P}$ be the distribution returned by Factor-Graph-Structure-Learn. Then for*

$$D(P\|\tilde{P}) + D(\tilde{P}\|P) \leq J\epsilon \quad (20)$$

---

[6]The upper bound is based on a very naive implementation running time.

*to hold with probability* $1 - \delta$, *it suffices that*

$$m \geq (1 + \tfrac{\epsilon \gamma^{k+b}}{2^{2k+3}})^2 \tfrac{v^{2k+2b} 2^{8k+19}}{\gamma^{6k+6b} \epsilon^4} \log \tfrac{8kbn^{k+b}v^{k+b}}{\delta}. \quad (21)$$

*Proof (sketch).* From Lemmas 9 and 10 we have that the conditioning set we pick gives a good approximation to the true canonical factor assuming we used true probabilities with that conditioning set. At this point the structure is fixed, and we can use the sample complexity theorem for parameter learning to finish the proof. □

Theorem 12 shows that the sample complexity depends exponentially on the maximum factor size $k$, the maximum Markov blanket size $b$, polynomially on $\frac{1}{\gamma}, \frac{1}{\epsilon}$. If we modify the analysis to consider the normalized KL-divergence, as in Section 3.4, we obtain a logarithmic dependence on the number of variables in the network.

To understand the implications of this theorem, consider the class of Gibbs distributions where every variable can participate in at most $d$ factors and every factor can have at most $k$ variables in its scope. Then we have that the Markov blanket size $b \leq dk^2$. Bayesian networks are also factor graphs. If the number of parents per variables is bounded by numP and the number of children is bounded by numC, then we have $k \leq \text{numP} + 1$, and that $b \leq (\text{numC}+1)(\text{numP}+1)^2$. Thus our factor graph structure learning algorithm allows us to efficiently learn distributions that can be represented by Bayesian networks with bounded number of children and parents per variable. Note that our algorithm recovers a distribution which is close to the true generating distribution, but the distribution it returns is encoded as a factor graph, which may not be representable as a compact Bayesian network.

Theorem 12 considers the case where the generating distribution $P$ actually factors according to a structure with size of factor scopes bounded by $k$ and size of factor Markov blankets bounded by $b$. As we did in the case of parameter estimation, we can show that we have graceful degradation of performance for distributions that do not satisfy this assumption.

**Theorem 13** (Structure learning: graceful degradation). *Let any* $\epsilon, \delta > 0$ *be given. Let* $\{\mathbf{x}^{(i)}\}_{i=1}^m$ *be IID samples from a distribution* $Q$. *Let* MB *and* $\widehat{\text{MB}}$ *be the Markov blankets according to the distributions* $Q$ *and found by* Factor-Graph-Structure-Learn *respectively. Let* $\{f^*_{\mathbf{D}^*_j | \text{MB}(\mathbf{D}^*_j)}\}_j$ *be the Markov blanket canonical factors of* $Q$. *Let* $\tilde{J}$ *be the number of factors in* $Q$ *with scopesize smaller than* $k$. *Let* $\tilde{P}$ *be the probability distribution returned by* Factor-Graph-Parameter-Learn. *Then we have that for*

$$D(Q \| \tilde{P}) + D(\tilde{P} \| Q) \leq \tilde{J}\epsilon + 2 \sum_{j : |\mathbf{D}^*_j| > k} \max_{\mathbf{d}^*_j} \left| \log f^*_{\mathbf{D}^*_j}(\mathbf{d}_j) \right|$$

$$+ 2 \sum_{j \,:\, |\mathbf{D}^*_j| \leq k, |\text{MB}(\mathbf{D}^*_j)| > b} \max_{\mathbf{d}^*_j} \left| \log \frac{f^*_{\mathbf{D}^*_j | \text{MB}(\mathbf{D}^*_j)}(\mathbf{d}^*_j)}{f^*_{\mathbf{D}^*_j | \widehat{MB}(\mathbf{D}^*_j)}(\mathbf{d}^*_j)} \right| \quad (22)$$

*to hold with probability at least* $1 - \delta$, *it suffices that* $m$ *satisfies Eqn. (21) of Theorem 12.*

Theorem 13 shows that (just like in the parameter learning setting) each canonical factor that is not captured by our learned structure contributes at most a constant to our bound on the KL-divergence. The reason a canonical factor is not captured could be two-fold. First, the scope of the factor could be too large. The paragraph after Theorem 5 discusses when the resulting error is expected to be small. Second, the Markov blanket of the factor could be too large. As shown in Eqn. (22), a good approximate Markov blanket is sufficient to get a good approximation. So we can expect these error contributions to be small if the true distribution is mostly determined by interactions between small sets of variables.

## 5 Related Work

### 5.1 Markov Networks

The most natural algorithm for parameter estimation in undirected graphical models is maximum likelihood (ML) estimation (possibly with some regularization). Unfortunately, evaluating the likelihood of such a model requires evaluating the partition function. As a consequence, all known ML algorithms for undirected graphical models are computationally tractable only for networks in which inference is computationally tractable. By contrast, our closed form solution can be efficiently computed from the data, even for Markov networks where inference is intractable. Note that our estimator does not return the ML solution, so that our result does not contradict the "hardness" of ML estimation. However, it does provide a low KL-divergence estimate of the probability distribution, with high probability, from a "small" number of samples, assuming the true distribution approximately factors according to the given structure.

Criteria different from ML have also been proposed for learning Markov networks. The most prominent is *pseudo-likelihood* (Besag, 1974), and its extension, generalized pseudo-likelihood (Huang & Ogata, 2002). The pseudo-likelihood criterion gives rise to a tractable convex optimization problem. However, the theoretical analyses (e.g., Geman & Graffigne, 1986; Comets, 1992; Guyon & Künsch, 1992) only apply when the generating model is in the true target class. Moreover, they show only asymptotic convergence rates, which are weaker than the finite sample size PAC-bounds we provide in our analysis. Pseudo-likelihood has been extended to obtain a consistent model selection procedure for a small set of models: the procedure is statistically consistent and an asymptotic convergence rate is given (Ji & Seymour, 1996). However, no algorithm is available to efficiently find the best pseudo-likelihood model over the exponentially large set of candidate models from which we want to select in the structure learning problem.

Structure learning for Markov networks is notoriously difficult, as it is generally based on using ML estimation of the parameters (with smoothing), often combined with a penalty term for structure complexity. As evaluating the likelihood is only possible for the class of Markov networks in which inference is tractable, there have been two main research tracks for ML structure learning. The first, starting with the work of Della Pietra et al. (1997), uses local-search heuristics to add factors into the network (see also McCallum, 2003). The second searches for a structure within a restricted class of models in which inference is tractable, more specifically, bounded treewidth Markov networks. Indeed, ML learning of the class of tree Markov networks — networks of tree-width 1 — can be performed very efficiently (Chow & Liu, 1968). Unfortunately, Srebro (2001) proves that for any tree-width $k$ greater than 1, even finding the ML treewidth-$k$ network is NP-hard. Srebro also provides an approximation algorithm, but the approximation factor is a very large multiplicative factor of the log-likelihood, and is therefore of limited practical use. Several heuristic algorithms to learn small-treewidth models have been proposed (Malvestuto, 1991; Bach & Jordan, 2002; Deshpande et al., 2001), but (not surprisingly, given the NP-hardness of the problem) they do not come with any performance guarantees.

Recently, Narasimhan and Bilmes (2004) provided a polynomial time algorithm with a polynomial sample complexity guarantee for the class of Markov networks of bounded tree width. They do not provide any graceful degradation guarantees when the generating distribution is not a member of the target class. Their algorithm computes approximate conditional independence information followed by dynamic programming to recover the bounded tree width structure. The parameters for the recovered bounded tree width model are estimated by standard ML methods. Our algorithm applies to a different family of distributions: factor graphs of bounded connectivity (including graphs in which inference is intractable). Factor graphs with small connectivity can have large tree width (e.g., grids) and factor graphs with small tree width can have large connectivity (e.g., star graphs).

### 5.2 Bayesian Networks

ML parameter learning in Bayesian networks (possibly with smoothing) only requires computing the empirical conditional probabilities of each variable given its parent instantiations.

Dasgupta (1997), following earlier work by Friedman and Yakhini (1996), analyzes the sample complexity of learning Bayesian networks, showing that the sample complexity is polynomial in the maximal number of different instantiations per family. His sample complexity result has logarithmic dependence on the number of variables in the network, when using the KL-divergence normalized by the number of variables in the network. In this paper, we strengthen his result, showing an $O(1)$ dependence of the number of samples on the number of variables in the network. So for bounded fan-in Bayesian networks, the sample complexity is independent of the number of variables in the network.

Results analyzing the complexity of structure learning of Bayesian networks fall largely into two classes. The first class of results assumes that the generating distribution is DAG-perfect with respect to some DAG $G$ with at most $k$ parents for each node. (That is, $P$ and $G$ satisfy precisely the same independence assertions.) In this case, algorithms based on various independence tests (Spirtes et al., 2000; Cheng et al., 2002) can identify the correct network structure at the infinite sample limit, using a polynomial number of independence tests. Chickering and Meek (2002) relax the assumption that the distribution be DAG-perfect; they show that, under a certain assumption, a simple greedy algorithm will, at the infinite sample limit, identify a network structure which is a minimal I-map of the distribution. They provide no polynomial time guarantees, but such guarantees might hold for models with bounded connectedness (such as the ones our algorithm considers).

The second class of results relates to the problem of finding a network structure whose score is high, for a given set of samples and some appropriate scoring function. Although finding the highest-scoring tree-structured network can be done in polynomial time (Chow & Liu, 1968), Chickering (1996) shows that the problem of finding the highest scoring Bayesian network where each variable has at most $k$ parents is NP-hard, for any $k \geq 2$. (See Chickering et al., 2003, for details.) Even finding the maximum likelihood structure among the class of polytrees (Dasgupta, 1999) and paths (Meek, 2001) is NP-hard. These results do not address the question of the number of samples for which the highest scoring network is guaranteed to be close to the true generating distribution.

Hoffgen (1993) analyzes the problem of PAC-learning the structure of Bayesian networks with bounded fan-in, showing that the sample complexity depends only logarithmically on the number of variables in the network (when considering KL-divergence normalized by the number of variables in the network). Hoffgen does not provide an efficient learning algorithm (and to date, no efficient learning algorithm is known), stating only that if the optimal network for a given data set can be found (e.g., by exhaustive enumeration), it will be close to optimal with high probability. By contrast, we provide a polynomial-time learning algorithm with similar performance guarantees for Bayesian networks with bounded fan-in and bounded fan-out. However, we note that our algorithm does not construct a Bayesian network representation, but rather a factor graph; this factor graph may not be compactly representable as a Bayesian network, but it is guaranteed to encode a distribution which is close to the generating distribution, with high probability.

## 6 Discussion

We have presented polynomial time algorithms for parameter estimation and structure learning in factor graphs of bounded factor size and bounded connectivity. When the generating distribution is within this class of networks, our algorithms are guranteed to return a distribution close to it, using a polynomial number of samples. When the generating distribution is not in this class, our algorithm degrades gracefully.

While of significant theoretical interest, our algorithms, as described, are probably impractical. From a statistical perspective, our algorithm is based on the canonical parameterization, which is evaluated relative to a canonical assignment $\bar{x}$. Many of the empirical estimates that we compute in the algorithm use only a subset of the samples that are (in some ways) consistent with $\bar{x}$. As a consequence, we make very inefficient use of data, in that many samples may never be used. In regimes where data is not abundant, this limitation may be quite significant in practice.

From a computational perspective, our algorithm uses exhaustive enumeration over all possible factors up to some size $k$, and over all possible Markov blankets up to size $b$. When we fix $k$ and $b$ to be constant, the complexity is polynomial. But in practice, the set of all subsets of size $k$ or $b$ is much too large to search exhaustively.

However, even aside from its theoretical implications, the algorithm we propose might provide insight into the development of new learning algorithms that do work well in practice. In particular, we might be able to address the statistical limitation by putting together canonical factor estimates from multiple canonical assignments $\bar{x}$. We might be able to address the computational limitation using a more clever (perhaps heuristic) algorithm for searching over subsets. Given the limitations of existing structure learning algorithms for undirected models, we believe that the techniques suggested by our theoretical analysis might be worth exploring.

**Acknowledgements.** This work was supported by DARPA contract number HR0011-04-1-0016.

## References

Abbeel, P., Koller, D., & Ng, A. Y. (2005). On the sample complexity of graphical models. (Full paper.) http://www.cs.stanford.edu/~pabbeel/.

Bach, F., & Jordan, M. (2002). Thin junction trees. *NIPS 14*.

Besag, J. (1974). Spatial interaction and the statistical analysis of lattice systems. *Journal of the Royal Statistical Society, Series B*.

Cheng, J., Greiner, R., Kelly, J., Bell, D., & Liu, W. (2002). Learning Bayesian networks from data: An information-theory based approach. *Artificial Intelligence Journal*.

Chickering, D. (1996). Learning Bayesian networks is NP-Complete. In D. Fisher and H. Lenz (Eds.), *Learning from data: Artificial intelligence and statistics v*, 121–130. Springer-Verlag.

Chickering, D., & Meek, C. (2002). Finding optimal Bayesian networks. *Proc. UAI*.

Chickering, D., Meek, C., & Heckerman, D. (2003). Large-sample learning of Bayesian networks is hard. *Proc. UAI*.

Chow, C. K., & Liu, C. N. (1968). Approximating discrete probability distributions with dependence trees. *IEEE Transactions on Information Theory*.

Comets, F. (1992). On consistency of a class of estimators for exponential families of Markov random fields on the lattice. *Annals of Statistics*.

Cover, T. M., & Thomas, J. A. (1991). *Elements of information theory*. Wiley.

Dasgupta, S. (1997). The sample complexity of learning fixed structure Bayesian networks. *Machine Learning*.

Dasgupta, S. (1999). Learning polytrees. *Proc. UAI*.

Della Pietra, S., Della Pietra, V. J., & Lafferty, J. D. (1997). Inducing features of random fields. *IEEE Transactions on Pattern Analysis and Machine Intelligence*, *19*, 380–393.

Deshpande, A., Garofalakis, M. N., & Jordan, M. I. (2001). Efficient stepwise selection in decomposable models. *Proc. UAI* (pp. 128–135). Morgan Kaufmann Publishers Inc.

Friedman, N., & Yakhini, Z. (1996). On the sample complexity of learning Bayesian networks. *Proc. UAI*.

Geman, S., & Graffigne, C. (1986). Markov random field image models and their applications to computer vision. *Proc. of the International Congress of Mathematicians*.

Guyon, X., & Künsch, H. (1992). Asymptotic comparison of estimators in the ising model. In *Stochastic models, statistical methods, and algorithms in image analysis, lecture notes in statistics*. Springer, Berlin.

Hammersley, J. M., & Clifford, P. (1971). Markov fields on finite graphs and lattices. Unpublished.

Hoffgen, K. L. (1993). Learning and robust learning of product distributions. *Proc. COLT*.

Huang, F., & Ogata, Y. (2002). Generalized pseudo-likelihood estimates for Markov random fields on lattice. *Annals of the Institute of Statistical Mathematics*.

Ji, C., & Seymour, L. (1996). A consistent model selection procedure for Markov random fields based on penalized pseudolikelihood. *Annals of Applied Probability*.

Kschischang, F. R., Frey, B. J., & Loeliger, H. A. (2001). Factor graphs and the sum-product algorithm. *IEEE Transactions on Information Theory*.

Lauritzen, S. L. (1996). *Graphical models*. Oxford University Press.

Malvestuto, F. M. (1991). Approximating discrete probability distributions with decomposable models. *IEEE Transactions on Systems, Man and Cybernetics*.

McCallum, A. (2003). Efficiently inducing features of conditional random fields. *Proc. UAI*.

Meek, C. (2001). Finding a path is harder than finding a tree. *Journal of Artificial Intelligence Research*, *15*, 383–389.

Narasimhan, M., & Bilmes, J. (2004). PAC-learning bounded tree-width graphical models. *Proc. UAI*.

Spirtes, P., Glymour, C., & Scheines, R. (2000). *Causation, prediction, and search (second edition)*. MIT Press.

Srebro, N. (2001). Maximum likelihood bounded tree-width Markov networks. *Proc. UAI*.